\documentclass[11pt]{article}

\usepackage{acl}
\usepackage{times}
\usepackage{float}
\usepackage{latexsym}
\usepackage[T1]{fontenc}
\usepackage[utf8]{inputenc}
\usepackage{microtype}
\usepackage{dblfloatfix}
\usepackage{inconsolata}
\usepackage{amsmath}
\usepackage{amssymb}
\usepackage{graphicx}
\usepackage{booktabs}
\usepackage{array}
\usepackage[dvipsnames]{xcolor}
\usepackage[normalem]{ulem}
\usepackage{cleveref}
\title{When Does Confidence Reveal Problem Difficulty?}

\author{
  Xiyu Bao \quad
  Langyuan Cui \quad
  Hwee Tou Ng \quad
  Chun Kai Ling \\
  Department of Computer Science\\
  National University of Singapore\\
  13 Computing Drive, Singapore 117417\\
  \texttt{\{e1718960,langyuan.c\}@u.nus.edu} \\
  \texttt{\{nght, chunkail\}@nus.edu.sg}
}

\newcommand{\pc}{\ensuremath{\mathrm{PC}}}

\begin{document}
\maketitle

\begin{abstract}
While confidence signals are commonly evaluated for their ability to predict a model's own correctness, their relationship with problem difficulty remains less understood. In this work, we investigate a wide range of confidence-related signals and whether they can serve as indicators of problem difficulty across various task domains. We find that perturbation inconsistency, semantic entropy, and solution length exhibit strong associations with population difficulty than token-level uncertainty. Notably, these relationships persist even when restricting the analysis to problems correctly answered by the focal model, suggesting that such signals capture information about difficulty beyond their ability to predict the model's own correctness. We further explore applications of these signals, illustrating the potential and limitations of using confidence-related signals to guide inference-time decisions. Our findings highlight the utility of model-specific signals as imperfect but informative proxies for population-level problem difficulty.
\end{abstract}

\section{Introduction}
Large language models (LLMs) do not perform uniformly across problems: they solve some reliably while others remain challenging even for more capable models. Two related notions are commonly used to characterize this variation. The first is \emph{model confidence}, which describes how certain a particular model is about its answer.
The second is \emph{item difficulty}, which instead characterizes how challenging a problem is. Although confidence and difficulty are related, they are fundamentally distinct. Confidence is specific to a model and its response, whereas difficulty captures the intrinsic property of the item relative to a solver population.

This relationship raises a natural question: \emph{can confidence signals themselves serve as indicators of item difficulty?} Difficulty estimates are useful for a variety of purposes. They allow benchmarks to characterize model performance beyond aggregate accuracy, identifying regions in which current models remain challenged \cite{ding2024easy2hard}. Difficulty information can also inform inference-time strategies, by adapting the amount or form of reasoning used for problems of different difficulty \cite{wang2025make}. However, obtaining reliable difficulty labels commonly requires responses from many human participants or models, which may be unavailable for newly constructed datasets. In contrast, confidence signals can often be elicited directly from a single LLM. If such signals reliably reflect item difficulty, they could provide a practical way to infer useful difficulty information without collecting a large population of responses.

Existing work has largely focused on evaluating confidence signals through their ability to discriminate correct from incorrect responses \cite{kadavath2022mostlyknow,kuhn2023semanticuncertainty,xiong2024can}. Meanwhile, work on item difficulty has typically obtained difficulty from aggregate human or model performance \cite{ding2024easy2hard}. Consequently, it remains unclear which, if any, commonly used confidence signals capture information about item difficulty, and whether this relationship generalizes across models and tasks.

In this work, we aim to study when confidence signals reveal not only whether a model is likely to be correct, but also something about the intrinsic challenge posed by the problem it is attempting to solve. Our contributions are as follows: we \textbf{(i)} formalize the distinction between model-specific confidence and population-level item difficulty and compare a range of commonly used confidence signals against difficulty estimates obtained from population response data, \textbf{(ii)} examine the generality of these relationships across tasks and models, \textbf{(iii)} analyse whether confidence signals continue to distinguish item difficulty when controlling for factors such as correctness, and finally \textbf{(iv)} examine two bounded inference-time applications: confidence-based model routing and cost-aware adaptation amongst reasoning procedures for a fixed model.

\section{Related Work}

\paragraph{Confidence Estimation in LLMs.}
A broad range of methods estimate the confidence or uncertainty of LLMs. These include elicited self-assessments in natural language \citep{kadavath2022mostlyknow,lin2022verbaluncertainty}, semantic uncertainty over multiple generations \citep{kuhn2023semanticuncertainty}, and direct confidence elicitation across tasks \citep{xiong2024can}. These approaches are predominantly evaluated by how well confidence predicts the correctness or factuality of the model's own response.
In contrast, we study whether these confidence signals also capture the difficulty of the underlying problem.

\paragraph{Item Difficulty Estimation.}
Item difficulty aims to characterise how challenging a problem is relative to a population of solvers. Item Response Theory (IRT), originally developed in psychometrics \citep{rasch1960probabilistic}, explicitly separates solver ability from latent item characteristics such as difficulty and discrimination. More recent benchmarks estimate continuous difficulty labels from large-scale human and LLM performance data \citep{ding2024easy2hard}, while other work explores answer-plausibility scores as a lower-cost alternative \citep{mozafari2026qdaps}. Population-based approaches require responses from many solvers; we instead ask whether confidence signals can provide a useful proxy for population-level item difficulty.
In contrast to these dedicated difficulty-estimation approaches, we ask whether confidence signals already commonly used to characterize LLM uncertainty can themselves serve as indicators of population-level item difficulty.

\paragraph{Connecting Confidence and Difficulty.}
Only limited work has explicitly examined their distinction. Some work adapts inference-time self-consistency using uncertainty as a proxy for problem difficulty \citep{wang2025make}. However, this line of work does not establish whether commonly used confidence signals themselves provide reliable estimates of population-level item difficulty.

\section{Difficulty and Confidence}
\label{sec:preliminaries}

\subsection{Problem Difficulty}

Let $\mathcal{P}$ denote a set of problems from a given task domain, such as mathematical reasoning or reading comprehension, where each problem $p=(q_p,a_p^*)\in\mathcal{P}$ consists of a question $q_p$ and a corresponding ground-truth answer $a_p^*$. Let $D(p)\in\mathbb{R}^+$ denote the latent difficulty of problem $p$, where larger values indicate greater difficulty. While difficulty may be conceptualized as an intrinsic property of a problem, its empirical measurement is generally relative to the capabilities of a population of answerers. In classical test theory, item difficulty is commonly assessed through the proportion of respondents who answer an item correctly \cite{lord2008statistical}. Similar performance-based approaches have been adopted for estimating problem difficulty in LLM benchmarks \cite{ding2024easy2hard}.
We therefore operationalize problem difficulty using the performance of a reference population $M$, a finite set of LLMs. Let $Y(m,p)\in\{0,1\}$ indicate whether model $m \in M$ correctly answers problem $p$, as determined by comparing its answer with the ground-truth answer $a_p^*$ under the task-specific correctness criterion. We estimate difficulty as the proportion of reference models that answer the problem incorrectly:
\begin{equation*}
    \widehat{D}_M(p)
    = \frac{1}{|M|}\sum_{m\in M}
    \mathbf{1}\!\left[Y(m,p)=0\right],
    \label{eq:population-difficulty}
\end{equation*}
where $\mathbf{1}[\cdot]$ denotes the indicator function. A higher value of $\widehat{D}_M(p)$ indicates that a larger fraction of reference models fail to solve $p$. Note that crucially, this measure is dependent on $M$.

\subsection{Model confidence.}
In contrast to difficulty, which characterizes a problem relative to a population of answerers, confidence is specific to an individual model and its response. We conceptualize confidence as a model's degree of certainty regarding the correctness of its generated answer, consistent with prior work on language model confidence estimation \cite{jiang2021know,kadavath2022mostlyknow}. Let $C(m,p)\in\mathbb{R}^+$ denote the latent confidence of model $m$ in its answer to problem $p$, with the dependence on the particular generated response left implicit for simplicity. 
Since confidence is not directly observable, prior work has proposed various observable signals as proxies for it \cite{xiong2024can,kadavath2022mostlyknow}. We refer to the model whose confidence is being examined as the \emph{focal model}. For a focal model $m$, let $S_k(m,p)\in\mathbb{R}^+$ denote the value of confidence-related signal $k$, computed from one or more outputs generated by model $m$ for problem $p$. Different signals capture different aspects of model behavior, such as predictive uncertainty, agreement across sampled responses, or stability under alternative formulations of the same problem. Consequently, they need not represent direct measurements of a single underlying internal belief.

\paragraph{Perturbation-based consistency.}
Perturbation consistency (PC) measures the stability of a model's answers under meaning-preserving reformulations of the same problem \citep{rabinovich2023semanticconsistency}. The underlying intuition is that a model that reliably solves a problem should produce consistent answers even when the problem is expressed differently. Conversely, variation across equivalent formulations may indicate uncertainty or sensitivity to the problem's presentation.
For each problem $(q_p,a^*_p)$, let $q_1,\ldots,q_K$ denote $K$ meaning-preserving perturbations, and let $\hat y_{m,i}$ be the normalized answer produced by model $m$ for perturbation $p_i$. We define PC as the proportion of answer pairs that agree on the same non-empty answer:
\begin{equation*}
    \begin{aligned}
    \pc(m,p)
    &= \frac{1}{\binom{K}{2}}
    \sum_{1\leq i<j\leq K} I_{ij},\\
    I_{ij}
    &= \mathbf{1}\!\left[
        \hat y_{m,i}=\hat y_{m,j}
        \;\wedge\;
        \hat y_{m,i}\neq\varnothing
    \right].
    \end{aligned}
    \label{eq:pc-definition}
\end{equation*}
Here, $\varnothing$ represents a response for which no valid answer can be extracted. Such responses are treated as non-matching, preventing missing answers from artificially increasing consistency.
We use $1-\pc(m,p)$ as our perturbation-inconsistency signal, such that larger values indicate greater answer instability. In our experiments, we use $K=7$ perturbations as described in Figure~\ref{fig:pc-example}, yielding $\binom{7}{2}=21$ pairwise comparisons per problem. We additionally consider a majority-pair variant, detailed in Appendix~\ref{app:signal-computation}.

\paragraph{Semantic answer entropy.}
Semantic entropy measures uncertainty across multiple sampled responses while accounting for the fact that distinct textual outputs may express the same answer \citep{kuhn2023semanticuncertainty}. Given a collection of responses to problem $p$, we group semantically equivalent answers into clusters and estimate the probability mass associated with each cluster. Entropy is then computed over the resulting distribution:
\begin{equation*}
    S_{\mathrm{SE}}(m,p)
    = -\sum_{c\in\mathcal{C}_{m,p}}
    \hat P(c\mid m,p)\log\hat P(c\mid m,p),
    \label{eq:semantic-entropy}
\end{equation*}
where $\mathcal{C}_{m,p}$ denotes the set of semantic answer clusters and $\hat P(c\mid m,p)$ their estimated probability masses. Higher entropy indicates that the model produces a greater diversity of semantically distinct answers, suggesting greater uncertainty. Unlike perturbation consistency, this signal characterizes variation across sampled responses rather than sensitivity to meaning-preserving changes in the input.

\paragraph{Prompted reports.}
Prompted reports estimate uncertainty or difficulty by explicitly asking the model to assess its own response or the problem being solved \citep{kadavath2022mostlyknow,lin2022verbaluncertainty,xiong2024can}. We consider three variants: (i) \emph{self-rated difficulty}, where the model assesses how challenging the problem is, (ii) \emph{numerical uncertainty}, where the model provides an explicit numerical assessment of its uncertainty ranging from 0 to 100, and (iii) \emph{verbalized uncertainty}, where the model explicitly reports its confidence using one of five categories: ``very sure'', ``sure'', ``neutral'', ``unsure'', or ``very unsure''. Unlike perturbation and sampling-based measures, these signals rely on the model's explicit self-assessment. Self-rated difficulty also differs conceptually from confidence estimation, as it directly asks the model to evaluate problem difficulty rather than its certainty.

\paragraph{Token-probability.}
Token-probability signals quantify uncertainty using the model's predictive distribution over next tokens \citep{malinin2021autoregressiveuncertainty,raj2022uncertaintysampling}. Let $P_m(v\mid p,y_{<t})$ denote the probability assigned by model $m$ to token $v$ at generation step $t$, conditioned on problem $p$ and previously generated tokens $y_{<t}$. We consider three measures:
(i) \emph{Mean token entropy}, which averages the entropy of the next-token distributions across generation steps, (ii) \emph{Least confidence}, which measures uncertainty based on the highest-probability token prediction, with lower maximum probabilities indicating greater uncertainty, and (iii) \emph{Margin uncertainty}, which measures ambiguity between the two most probable tokens, with smaller probability gaps indicating greater uncertainty.

\paragraph{Solution length.}
Finally, we consider solution length as an auxiliary behavioral indicator rather than a direct confidence measure. Let $T(m,p)$ denote the number of tokens generated by model $m$ when solving problem $p$. We define $S_{\mathrm{len}}(m,p)=T(m,p)$.
Longer completions may reflect greater reasoning effort or verbosity, and previous work has observed that models tend to generate longer solutions for harder HARP problems \citep{yue2024harp}. However, solution length is also influenced by model-specific generation behavior and does not necessarily indicate uncertainty.

Detailed definitions and implementations for all signals are provided in Appendix~\ref{app:signal-computation}. The prompt templates are given in Appendix~\ref{app:prompts}.

\begin{figure}[t]
    \centering
    \includegraphics[width=0.8\columnwidth]{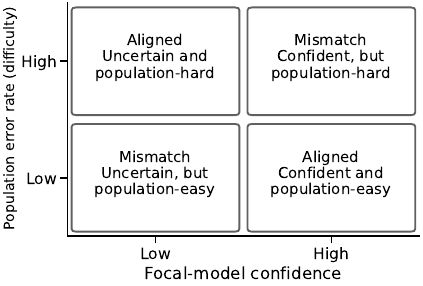}
    \caption{Conceptual distinction between model confidence and population difficulty. Aligned cases correspond to high confidence on relatively easy problems or low confidence on difficult problems. Mismatched cases illustrate that an individual model's confidence need not agree with the population difficulty.}
    \label{fig:confidence-difficulty-concept}
\end{figure}

\paragraph{Relationship between difficulty and confidence.}
Although model confidence and problem difficulty may be related, they represent distinct concepts. Population difficulty measures how challenging a problem is for a reference population, whereas confidence describes the certainty associated with a particular model's response. As illustrated in \Cref{fig:confidence-difficulty-concept}, a focal model may exhibit high confidence on a problem that is difficult for the reference population, or low confidence on one that is relatively easy. 
This distinction motivates our central research question: \emph{to what extent can confidence-related signals from an individual model serve as indicators of population-level problem difficulty?}

\begin{figure*}[t]
  \centering
  \small
  \setlength{\tabcolsep}{4pt}
  \renewcommand{\arraystretch}{1.12}
  \begin{tabular}{@{}p{0.17\textwidth}p{0.50\textwidth}cc@{}}
    \toprule
    Perturbation type   & Important transformed content (stored variant)                       & Parsed answer & Matches gold E? \\
    \midrule
    Stem rephrase       & ``$z$ is a complex number such that''; same equation and options     & E             & yes             \\
    Premise reorder     & Equation displayed before the question; options unchanged            & A             & no              \\
    Equivalent notation & Equivalent spacing/LaTeX notation for the same equation              & E             & yes             \\
    Structured outline  & ``Given'' / ``Constraints'' / ``Task'' / ``Answer Options'' sections & A             & no              \\
    Dialogue format     & ``Alice:'' / ``Bob:'' dialogue around the same equation              & $\varnothing$ & no              \\
    Option layout       & Answer choices rendered as a table                                   & A             & no              \\
    Translate Chinese   & Chinese translation of the stem and answer choices                   & $\varnothing$ & no              \\
    \bottomrule
  \end{tabular}
  \par\smallskip
  \parbox{0.98\textwidth}{\footnotesize All rows retain the same mathematical target: $12|z|^2 = 2|z+2|^2 + |z^2+1|^2 + 31$, with options $A=1$, $B=1/2$, $C=-1$, $D=4$, and $E=-2$. Matching pairs are $\binom{2}{2}+\binom{3}{2}=4$. The extracted answers therefore give $\pc=4/21=0.1905$ and $1-\pc=17/21=0.8095$.}
  \caption{An exact stored HARP/Qwen3-8B example (item \texttt{harp\_3818}). The original answer is correct and the population difficulty is $0.75$, but the seven meaning-preserving perturbations produce two \texttt{E}s, three \texttt{A}s, and two empty extractions. The resulting calculation is $\pc=4/21$ and $1-\pc=0.8095$.}
  \label{fig:pc-example}
\end{figure*}

\section{Experiments and Results}
\subsection{Experimental Setup}
\label{sec:experimental-setup}
\paragraph{Datasets and models}
We conduct our primary experiments on 500 items each from HARP \citep{yue2024harp} and E2H-AMC, the American Mathematics Competitions subset of Easy2Hard-Bench \citep{ding2024easy2hard}. Both benchmarks contain mathematical problems spanning a range of difficulties and provide human-derived difficulty information, making them suitable for investigating the relationship between model confidence and problem difficulty.
To estimate population-level difficulty, we construct a reference population of eight LLMs spanning different model families, sizes, and reasoning capabilities: Qwen3-8B \citep{qwen3report2025}, Phi-4-14B \citep{phi4report2024}, Gemma-3-12B and Gemma-3-27B \citep{gemma3report2025}, Llama-3.3-70B \citep{llama33modelcard2024}, Mistral Small 24B \citep{mistralsmall32025}, OLMo-2-13B \citep{olmo2furious2025}, and DeepSeek-R1-Distill-Llama-70B \citep{deepseekr12025}. This diversity aims to capture variation in model capabilities when estimating problem difficulty.

To examine whether the relationship between confidence signals and difficulty generalizes beyond mathematical reasoning, we extend our experiments to three additional benchmarks: SQuAD \citep{rajpurkar2016squad} for reading comprehension, ProofWriter \citep{tafjord2021proofwriter} for deductive reasoning, and MMLU \citep{hendrycks2021mmlu} for multidisciplinary knowledge and reasoning. We evaluate 500 items each from SQuAD and ProofWriter, and 600 MMLU items comprising 100 each from abstract algebra, college physics, econometrics, moral scenarios, professional medicine, and sociology. For these experiments, we use Phi-4-14B, Gemma-3-12B, and GPT-OSS-120B \citep{openai2025gptoss}, the models with usable runs across all three benchmarks.

\paragraph{Evaluation Protocol}
We extract the task-specific answer from each model's response and compare it against the dataset's reference answer to determine correctness. For each focal model $m$, we compute the population difficulty target $\widehat{D}_{M\setminus\{m\}}(p)$ using the remaining models in the reference population, following a leave-one-model-out protocol. This ensures that the focal model's own correctness does not directly contribute to the difficulty target against which its confidence signals are evaluated.
We assess the relationship between each signal $S_k(m,p)$ and population difficulty using Spearman's rank correlation coefficient. For consistency, we orient the signals such that larger values indicate greater uncertainty or predicted difficulty, so that positive correlations correspond to better alignment with population difficulty.

\subsection{Key Research Questions and Findings}

\paragraph{RQ1: Is model-population difficulty a meaningful measure?}
Before examining the relationship between confidence signals and problem difficulty, we first assess whether model-population error rate provides a meaningful proxy for difficulty. To this end, we compare the estimated difficulty $\widehat{D}_M(p)$ against the human-derived difficulty labels provided by HARP and E2H-AMC. For this analysis, we compute difficulty using the full reference population $M$, as no individual focal model is being evaluated. We observe positive Spearman correlations of $0.354$ ($95\%$ bootstrap CI: $[0.282, 0.430]$) on HARP and $0.535$ ($[0.467, 0.598]$) on E2H-AMC, indicating moderate agreement between model-population and human difficulty. These findings suggest that model-population error rate captures aspects of problem difficulty that are also reflected in human judgments, while the imperfect correlations highlight differences between human and model capabilities. We therefore adopt model-population error rate as our operational measure of difficulty in the subsequent analyses.

\paragraph{RQ2: Which signals correlate with population difficulty?}
We first examine the relationship between model-derived signals and population difficulty. Table~\ref{tab:raw-associations} reports the mean Spearman correlations across available model--dataset runs, with each run weighted equally. Perturbation inconsistency ($1-\pc$), semantic answer entropy, and solution length exhibit moderate positive correlations with difficulty on both datasets, whereas mean token entropy shows considerably weaker associations. In particular, $1-\pc$ achieves correlations of $0.557$ on E2H-AMC and $0.532$ on HARP, compared with $0.508$ and $0.509$ for semantic entropy, and $0.489$ and $0.573$ for solution length. In contrast, mean token entropy achieves correlations of only $0.262$ and $0.236$, respectively. These findings suggest that answer instability, semantic uncertainty, and solution length are more strongly associated with population difficulty than token-level uncertainty in the evaluated settings.

\begin{table}[t]
  \centering
  \small
  \begin{tabular}{@{}lrr@{}}
    \toprule
    Signal             & E2H-AMC & HARP  \\
    \midrule
    $1-\pc$            & 0.557   & 0.532 \\
    Semantic entropy   & 0.508   & 0.509 \\
    Solution length    & 0.489   & 0.573 \\
    Mean token entropy & 0.262   & 0.236 \\
    \bottomrule
  \end{tabular}
  \caption{Mean Spearman correlations between model-derived signals and population difficulty across available focal models. Each model--dataset run contributes equally to the average. Perturbation inconsistency ($1-\pc$) and semantic entropy are evaluated on four models per dataset, while solution length and mean token entropy are evaluated on eight.}
  \label{tab:raw-associations}
\end{table}

The associations are also consistent across models. Perturbation inconsistency exhibits positive correlations in all available runs, ranging from $0.489$ to $0.583$ on E2H-AMC and from $0.474$ to $0.585$ on HARP. Semantic entropy is also positively correlated with difficulty in all eight available runs. These results suggest that the observed relationships are not driven by a single focal model. 

\begin{figure*}[t]
  \centering
  \includegraphics[width=0.8\textwidth]{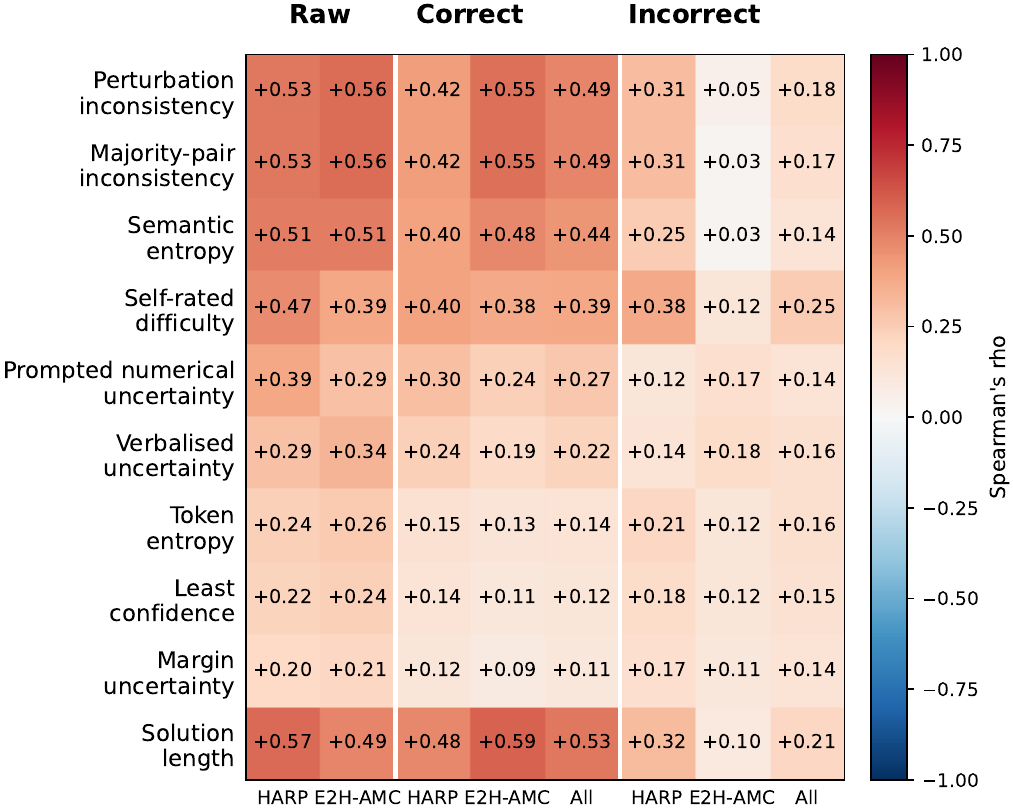}
  \caption{Associations between model-derived signals and population difficulty on E2H-AMC and HARP. Rows are signals and  while columns distinguish analyses over all problems, correctly answered problems, and incorrectly answered problems. Each cell is an equal-weight mean Spearman correlation.}
  \label{fig:all-signal-associations}
\end{figure*}

To provide a broader comparison, Figure~\ref{fig:all-signal-associations} summarizes the correlations for all ten core signals, including analyses restricted to correctly and incorrectly answered problems. Each cell represents the mean Spearman correlation across available focal models. Prompted reports and additional perturbation and token-based variants are presented in the appendix. Because signal availability differs across models, these results should be interpreted as an overview of observed associations rather than a definitive ranking of signal effectiveness.

\paragraph{RQ3: Do associations remain beyond correctness?}
By holding focal-model correctness constant, this analysis examines whether the signals can distinguish between problems that the focal model solves successfully but that differ in difficulty for the reference population. Table~\ref{tab:correct-associations} presents the resulting Spearman correlations, averaged equally across available focal models. Notably, perturbation inconsistency ($1-\pc$) remains positively associated with population difficulty. Similar positive associations are observed for semantic entropy and solution length, whereas mean token entropy exhibits a substantially weaker association. These findings suggest that answer instability, semantic uncertainty, and solution length contain information about population difficulty even among problems the focal model answers correctly.

\begin{table}[t]
  \centering
  \small
  \begin{tabular}{@{}lrrr@{}}
    \toprule
    Signal             & HARP  & E2H-AMC & All configs. \\
    \midrule
    $1-\pc$            & 0.423 & 0.550   & 0.486        \\
    Semantic entropy   & 0.398 & 0.476   & 0.437        \\
    Solution length    & 0.481 & 0.588   & 0.535        \\
    Mean token entropy & 0.150 & 0.133   & 0.142        \\
    \bottomrule
  \end{tabular}
  \caption{Mean Spearman correlations between model-derived signals and leave-one-model-out population difficulty, restricted to problems correctly answered by the focal model. Correlations are computed separately for each model--dataset configuration and averaged equally across available configurations.}
  \label{tab:correct-associations}
\end{table}

To further illustrate the distinction between answer correctness and perturbation consistency, Figure~\ref{fig:qualitative-regimes} presents four examples with Qwen3-8B as the focal model from the HARP dataset. These examples demonstrate that correctness and consistency need not coincide: a model may produce an incorrect answer despite highly consistent responses to perturbations, or correctly answer a problem while exhibiting substantial variation across perturbations. In particular, the latter case illustrates how perturbation inconsistency may reveal difficulty even when the focal model's original answer is correct. Together with the quantitative results, these examples suggest that perturbation consistency captures aspects of model behaviour not fully characterized by the correctness of a single response.


\begin{figure*}[t]
  \centering
  \small
  \setlength{\fboxsep}{5pt}
  \setlength{\fboxrule}{0.6pt}
  \begin{tabular}{@{}p{0.43\textwidth}@{\hspace{0.04\textwidth}}p{0.51\textwidth}@{}}
    
    \fcolorbox{black}{gray!7}{\begin{minipage}[t]{0.41\textwidth}
                                  \textbf{Easy $\mid$ correct $\mid$ consistent}\\[-1pt]
                                  \texttt{harp\_0}\\[2pt]
                                  \emph{Stem cue:} 64 split in the ratio 2:4:6; smallest part?\\
                                  \begin{tabular}{@{}ll@{}}
          Original / gold       & D / D                  \\
          Population difficulty & 0.125                  \\
          Seven perturbations   & \texttt{D D D D D D D} \\
          Inconsistency $1-\pc$ & 0.0000
        \end{tabular}
                                \end{minipage}}
     &
    \fcolorbox{black}{gray!7}{\begin{minipage}[t]{0.47\textwidth}
                                  \textbf{Easy $\mid$ incorrect $\mid$ consistent}\\[-1pt]
                                  \texttt{harp\_1008}\\[2pt]
                                  \emph{Stem cue:} $\sqrt{x+2}=2$; evaluate $(x+2)^2$.\\
                                  \begin{tabular}{@{}ll@{}}
          Original / gold       & D / B                  \\
          Population difficulty & 0.125                  \\
          Seven perturbations   & \texttt{B B B B B B B} \\
          Inconsistency $1-\pc$ & 0.0000
        \end{tabular}
                                \end{minipage}}
    \\[8pt]
    \fcolorbox{black}{gray!7}{\begin{minipage}[t]{0.41\textwidth}
                                  \textbf{Hard $\mid$ correct $\mid$ inconsistent}\\[-1pt]
                                  \texttt{harp\_3818}\\[2pt]
                                  \emph{Stem cue:} complex $z$ equation; evaluate $z+6/z$.\\
                                  \begin{tabular}{@{}ll@{}}
          Original / gold       & E / E                                                   \\
          Population difficulty & 0.750                                                   \\
          Seven perturbations   & \texttt{E A E A} $\varnothing$ \texttt{A} $\varnothing$ \\
          Inconsistency $1-\pc$ & 0.8095
        \end{tabular}
                                \end{minipage}}
     &
    \fcolorbox{black}{gray!7}{\begin{minipage}[t]{0.47\textwidth}
                                  \textbf{Hard $\mid$ incorrect $\mid$ inconsistent}\\[-1pt]
                                  \texttt{harp\_1513}\\[2pt]
                                  \emph{Stem cue:} checkerboard with a 1.5-inch card; maximum $n$?\\
                                  \begin{tabular}{@{}ll@{}}
          Original / gold       & A / B                  \\
          Population difficulty & 1.000                  \\
          Seven perturbations   & \texttt{B B A C A D D} \\
          Inconsistency $1-\pc$ & 0.8571
        \end{tabular}
                                \end{minipage}}
  \end{tabular}
  \caption{Examples from Qwen3-8B on HARP, showing combinations of population difficulty, focal-model correctness, and perturbation consistency. Each example reports the original and reference answers, population difficulty, answers to seven meaning-preserving perturbations, and perturbation inconsistency ($1-\pc$).}
  \label{fig:qualitative-regimes}
\end{figure*}

\paragraph{RQ4: Do the relationships generalize across model populations?}
Since population difficulty is defined relative to a set of reference models, its estimates may vary depending on the models selected. To assess this sensitivity, we partition the eight reference models into two disjoint groups of four and compare the resulting problem-difficulty rankings. The two groups exhibit Spearman correlations of $0.519$ on HARP and $0.849$ on E2H-AMC, indicating moderate agreement on HARP and substantially stronger agreement on E2H-AMC. These results suggest that the relative difficulty of problems is reasonably consistent across different model populations, particularly on E2H-AMC, although the choice of reference models continues to influence difficulty estimates. Results for additional population splits and individual models are provided in the appendix.

\begin{figure}[t]
  \centering
  \includegraphics[width=0.95\columnwidth]{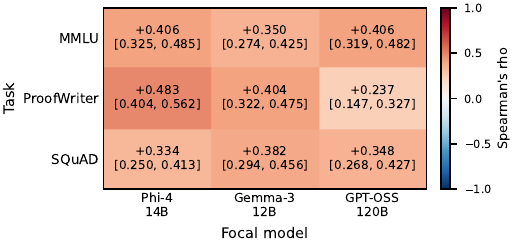}
  \caption{Spearman correlations between perturbation inconsistency ($1-\pc$) and population difficulty across MMLU, ProofWriter, and SQuAD. Each cell reports Spearman's $\rho$ with its $95\%$ confidence interval. 
  }
  \label{fig:cross-task-replication}
\end{figure}

\paragraph{RQ5: Do the relationships generalize across tasks?}
We further examine whether the observed relationship between perturbation inconsistency and difficulty extends beyond mathematical reasoning. As shown in Figure~\ref{fig:cross-task-replication}, perturbation inconsistency ($1-\pc$) remains positively correlated with population difficulty across all nine model--task configurations. Moreover, all corresponding $95\%$ confidence intervals exclude zero, providing further evidence that the relationship is not specific to mathematical problem solving.
The wider-domain experiments also evaluate token-probability signals, solution length, and prompted reports as auxiliary comparisons. We omit semantic entropy due to the additional sampling and semantic equivalence computations it requires. Furthermore, because models share evaluation items within each task and the confidence intervals are not adjusted for multiple comparisons, we interpret these results as a consistent pattern across domains rather than nine independent confirmations.

Overall, these results suggest that the relationship between perturbation inconsistency and population difficulty extends beyond the mathematics domain, while the stability of difficulty rankings depends on the composition of the reference population. Together with RQ2 and RQ3, our findings indicate that perturbation inconsistency provides information about the difficulty experienced by other models, including on problems the focal model answers correctly.

\section{Applications of Confidence Signals as Difficulty Predictors}
\label{sec:routing}

Having established that several confidence-related signals correlate with population difficulty, we further explore their potential practical utility. We stress that these experiments are intended as illustrative demonstrations of how difficulty-related signals may inform inference-time decisions, rather than attempts to develop state-of-the-art methods.

\subsection{Model Routing}

Routing aims to improve accuracy while limiting the number of queries to a stronger, more expensive model. Importantly, the most difficult problems are not necessarily the most beneficial to escalate, as even a stronger model may fail on them. An analysis of difficulty bands on E2H-AMC and MMLU shows that the largest accuracy gains from stronger-model escalation occur on intermediate-difficulty problems, as shown in Table~\ref{tab:double-threshold} This motivates an interval-based routing policy that targets problems exhibiting moderate uncertainty rather than simply escalating the most uncertain ones.

We evaluate Qwen3-8B and Phi-4-14B as weak models, with GPT-5.5 as the stronger model, across E2H-AMC, HARP, MMLU, and SQuAD. We use 30 calibration and 70 test items from each dataset, yielding 280 test items per weak model. Using perturbation consistency ($\pc$), we define a double-threshold policy:
\begin{equation*}
  \textsc{route}(p) =
  \mathbf{1}\!\left[
    \tau_1 \leq \pc(m,p) \leq \tau_2
    \right],
  \label{eq:double-threshold}
\end{equation*}
where thresholds are selected separately for each dataset using calibration items. Routed problems use the stronger model's answer, while all other problems retain the weak model's answer. We evaluate target routing budgets of 10\% and 20\%, comparing against one-sided perturbation-consistency thresholds, solution-length routing, token-entropy routing, and random routing.

Table~\ref{tab:double-threshold} summarizes the results. The interval policy improves accuracy over the strongest available comparator at each target budget, with gains ranging from $0.29$ to $4.36$ percentage points. The largest improvement occurs for Phi-4 at the 20\% budget, achieving $76.43\%$ accuracy compared with $72.07\%$ for random routing, while escalating only $13.57\%$ of test items. Although these findings suggest that perturbation consistency can inform selective escalation, the gains are modest in most settings and do not establish consistent superiority across models or tasks.
\begin{table}[t]
  \centering
  \small
  \begin{tabular}{@{}lrrrr@{}}
    \toprule
    Difficulty & $n$ & GPT-5.4 & Gemma/Phi avg. & $\Delta$ \\
    \midrule
    Hardest         & 33  & 0.333   & 0.000          & +0.333   \\
    Hard            & 35  & 0.686   & 0.229          & +0.457   \\
    Easy            & 37  & 0.919   & 0.608          & +0.311   \\
    Easiest         & 95  & 0.937   & 1.000          & $-0.063$ \\
    \bottomrule
  \end{tabular}
  \caption{Gain in performance from routing all problems from each difficulty from a weaker to a stronger model. $\Delta$ is GPT-5.4 accuracy minus the Gemma/Phi average.}
  \label{tab:e2h-routing-motivation}
\end{table}

\begin{table}[t]
  \centering
  \small
  \begin{tabular}{@{}lrrrr@{}}
    \toprule
    Model/budget    & Double-$\tau$ & Comparator     & Gain  \\
    \midrule
    Qwen3-8B / 10\% & 81.1         & Random (80.8) & +0.32    \\
    Qwen3-8B / 20\% & 82.1         & Random (81.9) & +0.29  \\
    Phi-4 / 10\%    & 71.1         & Length (70.4) & +0.71   \\
    Phi-4 / 20\%    & 76.4         & Random (72.1) & +4.36  \\
    \bottomrule
  \end{tabular}
  \caption{Test accuracy (\%) of the perturbation-consistency interval policy (Double-$\tau$) and the strongest available baseline at each routing budget. Gain is the accuracy improvement in percentage points, and Routed is the actual percentage of test items escalated. Results are aggregated over 280 test items per weak model.}
  \label{tab:double-threshold}
\end{table}

\subsection{Adaptive Inference}

We further investigate whether observable signals can guide the allocation of reasoning effort within a fixed model. Using Qwen3-8B and Phi-4-14B on the same four tasks, we consider three reasoning strategies: chain-of-thought (CoT) \citep{wei2022chain}, self-refinement \citep{madaan2023selfrefine}, and least-to-most decomposition \citep{zhou2023leasttomost}. We evaluate an offline threshold policy that selects a reasoning strategy using signals from an initial CoT completion, including token entropy, token-probability margins, and solution length. Thresholds are selected using calibration data, with full implementation details provided in Appendix~\ref{app:adaptive-inference}.

For Qwen3-8B, the adaptive policy achieves $85.58\%$ accuracy at an average of 6,121 generated tokens per item, compared with $85.83\%$ for CoT alone and $86.67\%$ for always-on self-refinement. For Phi-4-14B, it achieves $72.83\%$ accuracy at 1,359 tokens per item, compared with $72.92\%$ for CoT alone. These results illustrate the potential for signal-guided inference allocation, although the adaptive policies do not consistently improve accuracy over fixed reasoning strategies. We therefore view these experiments as preliminary evidence of possible cost--quality trade-offs rather than a demonstration of superior adaptive inference.

\section{Conclusion}
In this work, we investigated whether confidence-related signals from LLMs can serve as indicators of problem difficulty. Across various task domains, we found that several signals exhibit positive associations with population difficulty. While the strength of these relationships varies across signals, models, and tasks, our findings highlight the potential of model-specific signals for estimating population-level difficulty. We further explored their practical utility in model routing and adaptive inference, demonstrating opportunities for selective escalation. Overall, our results suggest that confidence-related signals offer useful, though imperfect, proxies for problem difficulty.

\section{Limitations}
Our difficulty target is defined by a finite, task-dependent reference population, so it should not be interpreted as intrinsic or human difficulty. We do not have every signal for every model and task, so some comparisons use different sets of runs. The routing and fixed-model studies are offline evaluations on held-out subsets with modest sample sizes; their results and thresholds may not transfer to new models, prompts, or tasks. Finally, perturbation-based signals require extra generations and therefore have a meaningful deployment cost, while the auxiliary prompted and token-level measures were not evaluated for every configuration.

\paragraph{Responsible use and potential impacts.} Our signals are research diagnostics, not reliable measures of intrinsic difficulty, human ability, or model safety. Using them to route, rank, or deprioritise inputs could systematically misallocate computation when performance differs across task domains, languages, populations, or model families. Incorrect routing may deny stronger-model processing to an item that would benefit from it, or needlessly escalate an item with sensitive content. We therefore recommend use only with task-specific validation, calibration on representative data, monitoring for differential failure patterns, and human oversight in consequential settings. The present work is an offline evaluation and does not support deployment claims.

\bibliography{references}

\clearpage
\appendix
\section{Implementation Details}
\label{app:implementation}

This section describes the data, prompts, computation, and evaluation protocols used
to obtain the reported results. 

\subsection{Data, models, and computational environment}
\label{app:data-environment}

\paragraph{Data partitioning.} The primary analysis uses the 500-item subsets of HARP and E2H-AMC. The
wider-domain analysis uses 500 SQuAD items, 500 ProofWriter items, and 600 MMLU items distributed equally across the six reported subjects.
\paragraph{Model populations.} The reference population contains eight models for the primary analysis. The
wider-domain analysis uses Phi-4-14B, Gemma-3-12B, and GPT-OSS-120B because these
models have complete usable coverage across the three additional tasks.
The routing and adaptive-inference use the held-out partitions described in
Section~\ref{app:protocols}.
\paragraph{Computational environment.} Local model inference was served with vLLM on four NVIDIA A100-SXM4 GPUs with 40\,GB
of memory each. Hosted GPT-5.5 calls were treated as a separate service component and
were not assigned a local GPU cost. The analysis pipeline is Python-based and uses the
persisted run configurations to record model identifiers, decoding settings, token
counts, finish reasons, and parsed answers.
\paragraph{Data and artefact handling.} All datasets and model checkpoints are existing research artefacts; their creators and
source papers are cited in the main text. We collect no new human data or personal information, and the paper redistributes neither model weights nor benchmark content beyond the derived statistics needed for the analysis.

\paragraph{Reproducibility and compute reporting.} The environment specifies Python 3.11, PyTorch with CUDA 12.1, vLLM version 0.6.4.post1 or later, pandas version 2.0 or later, SciPy version 1.11 or later, Transformers version 4.30 or later, Accelerate version 0.20 or later, Hugging Face Hub version 0.20 or later, Datasets version 2.14 or later, tqdm version 4.65 or later, python-dotenv version 1.0 or later, OpenAI version 2.45.0, and four NVIDIA A100-SXM4-40GB GPUs for the documented local environment. Primary local runs use 8,192-token output limits; wider-domain runs use 1,024-token limits; perturbation generation uses 1,024-token limits and eight-token validation limits; Qwen3 adaptive inference uses 16,384, 8,192, and 4,096-token stage limits with a 32,768-token context; Phi-4 adaptive inference uses 6,144, 6,144, 3,072, and 8,192-token stage limits with a 16,384-token context. No model hyperparameter search was performed; routing thresholds were selected on calibration data and adaptive policies were selected on validation data before frozen test evaluation.

\paragraph{Statistical reporting.} The reported bootstrap intervals use 1,000 replicates and the 2.5th and 97.5th percentile boundaries, with resampling clustered by problem ID where multiple model responses share an item. Aggregate tables compute panel-level correlations before equal-weight averaging across available model--dataset configurations.

\paragraph{Use of AI assistants.} AI assistants were used only for debugging and code organisation.

\subsection{Dataset provenance, selection, and artifact use}
\label{app:datasets-artifacts}

For the primary analysis, HARP items are loaded from the public HARP multiple-choice
release and E2H-AMC items from the Easy2Hard-Bench AMC release. The loader reconstructs
each question from its canonical stem and answer choices where needed, retains the
published answer and difficulty field, removes empty questions, and assigns stable item
IDs. When a 500-item analysis subset is requested, items are ordered by the available
difficulty field and sampled at evenly spaced positions. This preserves coverage across
the difficulty range instead of selecting the first 500 records.

The wider-domain runs use the SQuAD validation split, the ProofWriter test split, and the
MMLU test split. MMLU contributes 100 items from each of the six subjects named in the
main text. SQuAD prompts retain the supplied context and question, ProofWriter prompts
retain the theory and query, and MMLU prompts retain all four choices and their labels.
Prompt construction requires these fields; rows that cannot be normalized into the
required format are not used. The routing and adaptive-inference studies use frozen
manifests so that
calibration, validation, and test items are selected before model outputs are collected;
their test partitions are not used for threshold fitting.

The study does not conduct a human-participant experiment, so participant instructions,
consent, compensation, demographics, and ethics-board procedures are not applicable.
The code record the dataset identifiers, item IDs,
model identifiers, prompt settings, decoding parameters, token counts, and parsed
outputs needed to audit the reported analyses. We use upstream datasets, checkpoints,
and hosted services under their respective terms and do not redistribute benchmark text,
model weights, or other restricted artifacts.

\subsection{Prompt templates and decoding}
\label{app:prompts}
\newcommand{\promptnl}{\newline}
\newcommand{\promptrow}[2]{#1 & \begin{minipage}[t]{0.76\textwidth}\raggedright #2\end{minipage}\\[5pt]}
\newcommand{\prompttableformat}{\footnotesize\renewcommand{\arraystretch}{1.18}}

The following are the prompts used by the experiments reported in the paper. We write
``\texttt{<item>}'' for the inserted problem and ``\texttt{<question>}'' for the
question field of a generated perturbation. 

The tables below show the complete prompt templates used by the reported experiments.
The left column identifies the setting and the right column gives the prompt text; angle
brackets denote fields filled with the current item or generated content.

\begin{table*}[t]
\centering\prompttableformat
\begin{tabular}{@{}>{\raggedright\arraybackslash}p{0.18\textwidth}>{\raggedright\arraybackslash}p{0.76\textwidth}@{}}
\toprule
\textbf{Setting} & \textbf{Prompt text} \\
\midrule
\promptrow{HARP original system}{You are an expert problem solver. Solve the multiple-choice problem and give your final answer as a single letter. End with `Answer: <letter>'.}
\promptrow{E2H-AMC original system}{You are a mathematics expert. Solve the problem and put your final answer inside \texttt{\\textbackslash boxed\{\}}.}
\promptrow{Original user message}{The canonical benchmark item, inserted verbatim.}
\promptrow{Perturbation generator system}{You are an expert problem editor. Follow the instructions exactly and output only the rewritten problem, with no preamble or explanation.}
\promptrow{Perturbation suffix}{Preserve every fact, logical condition, mathematical value, variable identity, unit, required output, answer option, option label, and which option or value would be correct. Do not add, omit, infer, or alter information. Do not solve the problem, reveal the correct answer, add a solution, or add commentary. The rewrite must be visibly different from the original; do not return an identical copy. Output ONLY the rewritten problem. Problem: \texttt{<question>}}
\bottomrule
\end{tabular}
\caption{Primary original-response and perturbation-generation prompts. Each operation below was concatenated with the generator system message and the common suffix.}
\label{tab:primary-prompts}
\end{table*}

\begin{table*}[t]
\centering\prompttableformat
\begin{tabular}{@{}>{\raggedright\arraybackslash}p{0.18\textwidth}>{\raggedright\arraybackslash}p{0.76\textwidth}@{}}
\toprule
\textbf{Setting} & \textbf{Prompt text} \\
\midrule
\promptrow{HARP: stem rephrase}{Rewrite the problem stem using substantially different synonymous wording. Keep every answer option and its label verbatim.}
\promptrow{HARP: premise reorder}{Change the order in which the givens and conditions are presented. Keep the final question and every answer option verbatim.}
\promptrow{HARP: equivalent notation}{Replace mathematical expressions or conditions with equivalent notation without changing their values or domains. Keep every answer option and its label verbatim.}
\promptrow{HARP: structured outline}{Convert the prose stem into a structured Given / Constraints / Task outline. Keep every answer option and its label verbatim.}
\promptrow{HARP: dialogue}{Present the stem as a concise dialogue that divides the givens, constraints, and final question between speakers. Keep every answer option and its label verbatim.}
\promptrow{HARP: option layout}{Keep the stem and option text verbatim, but present the labelled answer options in a visually different layout such as a compact table. Keep their original order and labels.}
\promptrow{HARP: Chinese translation}{Translate all prose into Chinese while keeping every number, symbol, formula, answer-option label, and option order unchanged.}
\promptrow{E2H-AMC: variable rename}{Consistently rename only abstract variables, points, sets, indices, or symbolic labels that are present in the problem. If the problem has no such labels, rephrase named quantities without changing names, numbers, or relationships.}
\promptrow{E2H-AMC: premise reorder}{Change the presentation order of independent givens and conditions only. Preserve chronological, causal, and procedural order exactly when order affects the mathematics. Keep the final question last.}
\promptrow{E2H-AMC: equivalent notation}{Replace mathematical expressions with exactly equivalent notation or representations without changing values or domains.}
\promptrow{E2H-AMC: condition rewrite}{Rewrite conditions as logically equivalent statements, such as a congruence as a remainder statement or an interval as inequalities. At least one condition must be rewritten; do not return the original wording unchanged.}
\promptrow{E2H-AMC: structured outline}{Convert the prose problem into a structured Given / Constraints / Find outline.}
\promptrow{E2H-AMC: dialogue}{Present the givens, constraints, and final question as a concise dialogue between two speakers. The dialogue must ask the same question but must not include a worked solution, intermediate calculations, hints, or the final answer.}
\promptrow{E2H-AMC: Chinese translation}{Translate all prose into Chinese while keeping every number, symbol, formula, and variable unchanged.}
\bottomrule
\end{tabular}
\caption{Operation prompts used to generate the primary HARP and E2H-AMC variants.}
\label{tab:primary-operations}
\end{table*}

\begin{table*}[t]
\centering\prompttableformat
\begin{tabular}{@{}>{\raggedright\arraybackslash}p{0.18\textwidth}>{\raggedright\arraybackslash}p{0.76\textwidth}@{}}
\toprule
\textbf{Setting} & \textbf{Prompt text} \\
\midrule
\promptrow{Equivalence validator system}{You are a rigorous equivalence validator.}
\promptrow{Equivalence validator user}{Below are two versions of the same problem. Decide whether they are logically equivalent --- i.e. they pose the same task and must yield the same correct answer, even if phrased, formatted, or translated differently. The modified version may be in Chinese. Reply NO if the meaning or identity of any fact, condition, mathematical value, variable, unit, answer option, option label, or required output was added, omitted, or changed. Reply NO if the modified version is an identical copy, reveals the answer, gives a solution, adds hints, or changes the order of time-dependent or procedural events. Original: \texttt{<original>} Modified: \texttt{<modified>} Are these two problems logically equivalent? Reply with exactly one word: YES or NO.}
\promptrow{Self-rating system}{You are an expert problem difficulty grader.}
\promptrow{Self-rating 1}{Rate how hard the following problem is for a strong AI model on an integer scale from 1 to 5, using this rubric: 1 = trivial, solvable instantly with no reasoning; 2 = easy, a single routine step; 3 = moderate, a few reasoning steps; 4 = hard, multi-step or non-obvious insight required; 5 = very hard, competition / olympiad level. Reply with ONLY the single digit (1-5). Do not explain.\promptnl\promptnl Problem: \texttt{<item>}}
\promptrow{Self-rating 2}{Do not solve the problem. Return exactly one digit from 1 to 5 for how hard it is for a strong AI model.\promptnl\promptnl Problem: \texttt{<item>}}
\promptrow{Self-rating 3}{Output format: <digit>. Allowed digits: 1, 2, 3, 4, 5. No words.\promptnl\promptnl Problem: \texttt{<item>}}
\promptrow{Uncertainty system}{You are a calibrated problem-solving confidence estimator.}
\promptrow{Numerical uncertainty 1}{Estimate the probability that you would answer the problem below correctly. Reply with ONLY one integer from 0 to 100, where 0 means certainly incorrect and 100 means certainly correct.\promptnl\promptnl Problem: \texttt{<item>}}
\promptrow{Numerical uncertainty 2}{Do not solve the problem. Give only your calibrated confidence as a percentage integer from 0 to 100.\promptnl\promptnl Problem: \texttt{<item>}}
\promptrow{Numerical uncertainty 3}{Output format: Confidence: <integer 0-100>. No reasoning. No answer choice. Estimate only.\promptnl\promptnl Problem: \texttt{<item>}}
\promptrow{Verbal uncertainty 1}{Rate how certain you are that you would answer the problem below correctly. Reply with ONLY one label: very sure, somewhat sure, uncertain, somewhat unsure, very unsure.\promptnl\promptnl Problem: \texttt{<item>}}
\promptrow{Verbal uncertainty 2}{Do not solve the problem. Choose exactly one certainty label from this list: very sure, somewhat sure, uncertain, somewhat unsure, very unsure.\promptnl\promptnl Problem: \texttt{<item>}}
\promptrow{Verbal uncertainty 3}{Output format: Certainty: <one of very sure, somewhat sure, uncertain, somewhat unsure, very unsure>. No reasoning.\promptnl\promptnl Problem: \texttt{<item>}}
\promptrow{Semantic-equivalence system}{You are a rigorous answer-equivalence judge.}
\promptrow{Semantic-equivalence user}{Decide whether the two candidate answers below mean the same thing as answers to the given problem. Consider mathematically equivalent forms identical. Reply with exactly YES or NO.\promptnl\promptnl Problem: \texttt{<item>}\promptnl\promptnl Answer 1:\promptnl\texttt{<first>}\promptnl\promptnl Answer 2:\promptnl\texttt{<second>}}
\bottomrule
\end{tabular}
\caption{Validator and prompted-confidence prompts, including all retry variants used in the reported baselines.}
\label{tab:baseline-prompts}
\end{table*}

\begin{table*}[t]
\centering\prompttableformat
\begin{tabular}{@{}>{\raggedright\arraybackslash}p{0.18\textwidth}>{\raggedright\arraybackslash}p{0.76\textwidth}@{}}
\toprule
\textbf{Setting} & \textbf{Prompt text} \\
\midrule
\promptrow{SQuAD original}{System: Answer the reading-comprehension item. Reason briefly if needed, then end with a line exactly in the form `Answer: <short span>'. User: Answer the question using a short span copied from the context. Context: \texttt{<context>} Question: \texttt{<question>} Answer:}
\promptrow{ProofWriter original}{System: Solve the logical proof item. Reason briefly if needed, then end with a line exactly in the form `Label: True', `Label: False', or `Label: Unknown'. User: Given the theory, answer whether the query is True, False, or Unknown. Return only one label. Theory: \texttt{<theory>} Query: \texttt{<question>} Label:}
\promptrow{MMLU original}{System: Answer the multiple-choice question. Reason briefly if needed, then end with a line exactly in the form `Answer: A', `Answer: B', `Answer: C', or `Answer: D'. User: Answer the multiple-choice question. Return only the letter A, B, C, or D. Question: \texttt{<question>} A. \texttt{<choice A>} B. \texttt{<choice B>} C. \texttt{<choice C>} D. \texttt{<choice D>} Answer:}
\promptrow{Wider-domain generator}{You rewrite evaluation items while preserving the correct answer.}
\promptrow{SQuAD preservation}{Preserve the exact requested answer span, entities, numbers, reference, and required output. Do not add hints, new facts, or answer-revealing text. Output only the rewritten question.}
\promptrow{ProofWriter theory preservation}{Preserve every fact, rule, entity, predicate, negation, quantifier, implication, and dependency. Do not add hints, new facts, stronger assumptions, weaker assumptions, answers, labels, or answer-revealing text. Output only the rewritten theory.}
\promptrow{ProofWriter query preservation}{Preserve the exact query meaning, entity, predicate, and negation. Do not add hints, new facts, labels, answers, or answer-revealing text. Output only the rewritten query.}
\promptrow{MMLU preservation}{Preserve the exact meaning, task, answer set, correct answer, entities, numbers, labels, logical relations, and required output. Do not add hints, new facts, stronger assumptions, weaker assumptions, or answer-revealing text. Output only the rewritten item.}
\bottomrule
\end{tabular}
\caption{Original-response and preservation prompts for the wider-domain experiments.}
\label{tab:wider-original-prompts}
\end{table*}

\begin{table*}[t]
\centering\prompttableformat
\begin{tabular}{@{}>{\raggedright\arraybackslash}p{0.18\textwidth}>{\raggedright\arraybackslash}p{0.76\textwidth}@{}}
\toprule
\textbf{Setting} & \textbf{Prompt text} \\
\midrule
\promptrow{SQuAD: rephrase}{Rewrite the question using different natural wording.}
\promptrow{SQuAD: relation explicit}{Rewrite the question so it makes the searched relation explicit, but still asks for the same short answer span.}
\promptrow{SQuAD: references explicit}{Rewrite the question by making pronoun or definite-description references more explicit when the passage context permits it.}
\promptrow{SQuAD: answer type}{Rewrite the question so the requested answer type is clearer near the start of the question, without changing the requested span.}
\promptrow{SQuAD: grammar}{Rewrite the question with a different grammatical structure, such as active to passive voice or clause reordering, while preserving meaning.}
\promptrow{SQuAD: question form}{Rewrite the question using an equivalent who, what, when, where, which, or how form where appropriate.}
\promptrow{SQuAD: concise}{Rewrite the question more concisely while keeping all constraints needed to identify the same answer span.}
\promptrow{ProofWriter: reorder}{Reorder independent facts and rules in the theory while preserving every fact, rule, negation, quantifier, and implication.}
\promptrow{ProofWriter: query rephrase}{Rewrite only the query sentence using equivalent wording. Keep the theory verbatim.}
\promptrow{ProofWriter: sections}{Present the theory as Facts and Rules sections while preserving each statement exactly in meaning.}
\promptrow{ProofWriter: implications}{Rewrite implication rules with equivalent if/then wording. Do not alter antecedents, consequents, negations, conjunctions, or entity names.}
\promptrow{ProofWriter: negation}{Rewrite statements containing negation into an equivalent explicit form such as `it is not the case that'. Do not add or remove any negation, rule condition, exception, or entity.}
\promptrow{ProofWriter: grouping}{Group facts and rules by the predicates they mention when this is possible without changing the proof problem. Keep every dependency intact.}
\promptrow{ProofWriter: formatting}{Change punctuation, line breaks, and spacing to improve readability. Do not change any token that carries logical meaning.}
\promptrow{MMLU: stem rephrase}{Rewrite only the question stem using equivalent wording. Keep all four choices and their labels verbatim and in the same order.}
\promptrow{MMLU: option layout}{Keep the question stem and all choice text verbatim, but present the labelled choices in a different clear layout. Preserve labels A-D and their order.}
\promptrow{MMLU: subject header}{Add only a brief neutral subject header matching the field of the question. Do not add facts or hints. Keep the stem and choices semantically unchanged.}
\promptrow{MMLU: terminology}{Replace non-essential wording in the stem with synonymous terminology that is standard for the same subject. Keep choices verbatim.}
\promptrow{MMLU: exam record}{Present the item as an exam record with Question and Options fields. Keep all answer options, labels, order, and the correct answer unchanged.}
\promptrow{MMLU: copy-edit}{Lightly copy-edit the stem for readability without changing any technical condition, named entity, quantity, or asked-for relation. Keep choices verbatim.}
\promptrow{MMLU: answer instruction}{Rewrite the item so the instruction to choose the best answer appears before the question stem. Keep the stem meaning, all choices, labels, choice order, and correct answer unchanged.}
\promptrow{Wider wrappers}{SQuAD: `The passage below is read-only. Do not rewrite, summarise, reorder, or quote the passage in your output. Return only one rewritten question. Passage: <context> Original question: <question> Rewritten question:'. ProofWriter query: `The theory below is read-only. Do not rewrite or quote the theory in your output. Return only one rewritten query sentence, with no label and no explanation. Theory: <theory> Query: <question> Rewritten query:'. ProofWriter theory: `The query below is read-only. Do not rewrite or answer it. Return only the rewritten theory, preserving every fact, rule, entity, negation, and implication. Do not include a Label field or an answer. Theory: <theory> Query: <question> Rewritten theory:'. MMLU: `Rewrite this complete evaluation item. Keep any answer cue such as Answer: or Label: as an empty output field; do not include the correct answer. Item: <item>'.}
\promptrow{SQuAD validator}{Decide whether these two SQuAD questions ask for exactly the same answer span from the same passage. Reply NO if the modified question adds a hint, changes the entity, changes the requested relation, or could select a different span. Original question: \texttt{<original-question>} Modified question: \texttt{<modified-question>} Are they equivalent? Reply with exactly one word: YES or NO.}
\promptrow{ProofWriter validator}{Decide whether these two ProofWriter items define exactly the same logic problem and therefore must have the same True, False, or Unknown label. Reply NO if any fact, rule, entity, predicate, negation, implication, or query meaning was added, omitted, weakened, or changed. No omitted field is allowed. Original and modified theory/query fields are provided, followed by: Are they equivalent? Reply with exactly one word: YES or NO.}
\promptrow{MMLU validator}{Decide whether the original and modified items are equivalent. Reply with YES or NO. The original and modified items are then provided.}
\bottomrule
\end{tabular}
\caption{Operation prompts and wrappers for the SQuAD, ProofWriter, and MMLU perturbation experiments.}
\label{tab:wider-operations}
\end{table*}

\begin{table*}[t]
\centering\prompttableformat
\begin{tabular}{@{}>{\raggedright\arraybackslash}p{0.18\textwidth}>{\raggedright\arraybackslash}p{0.76\textwidth}@{}}
\toprule
\textbf{Setting} & \textbf{Prompt text} \\
\midrule
\promptrow{Routing systems}{HARP: You are an expert problem solver. End with `Answer: <letter>'. E2H-AMC: You are a mathematics expert. Put your final answer in \texttt{\\textbackslash boxed\{\}}. MMLU: Return only the letter A, B, C, or D. SQuAD: Return only a short answer span from the supplied context.}
\promptrow{Routing: original}{\texttt{<item>}}
\promptrow{Routing: direct}{Answer the following item.\promptnl\promptnl\texttt{<item>}}
\promptrow{Routing: careful}{Read the following item carefully, then answer it.\promptnl\promptnl\texttt{<item>}}
\promptrow{Routing: step-by-step}{Solve the following item step by step before giving the requested final answer.\promptnl\promptnl\texttt{<item>}}
\promptrow{Routing: verify}{Solve the following item, check your reasoning once, and then give the requested final answer.\promptnl\promptnl\texttt{<item>}}
\promptrow{Routing: independent}{Treat the following as a self-contained item and solve it independently.\promptnl\promptnl\texttt{<item>}}
\promptrow{Routing: concise}{Answer the following item efficiently and follow its requested answer format.\promptnl\promptnl\texttt{<item>}}
\bottomrule
\end{tabular}
\caption{Prompt-only routing templates. The original item is inserted verbatim in every user prompt.}
\label{tab:routing-prompts}
\end{table*}

\begin{table*}[t]
\centering\scriptsize
\begin{tabular}{@{}>{\raggedright\arraybackslash}p{0.18\textwidth}>{\raggedright\arraybackslash}p{0.76\textwidth}@{}}
\toprule
\textbf{Setting} & \textbf{Prompt text} \\
\midrule
\promptrow{Final-answer contracts}{HARP: After all reasoning, make the final visible line exactly `Final answer: X', where X is A, B, C, D, or E. Do not add explanation after that line. MMLU uses A--D; E2H-AMC uses `Final answer: \texttt{\\textbackslash boxed\{...\}}'; SQuAD uses `Final answer: <answer>' with the shortest supported span and no explanation afterward.}
\promptrow{Chain-of-thought system}{Solve the problem carefully. Work through the reasoning step by step, check the result, and then give the final answer, followed by the applicable final-answer contract.}
\promptrow{Refinement system}{Act as a meticulous solver revising an earlier attempt. Independently verify the interpretation, evidence, calculations, and answer choice; correct every error you find. Give a coherent revised solution, not merely a critique, followed by the applicable contract.}
\promptrow{Refinement user}{Original problem: \texttt{<problem>} Earlier chain-of-thought attempt (untrusted; it may contain errors): \texttt{<draft>} Produce the corrected solution now.}
\promptrow{Decomposition system}{Decompose the problem into the smallest useful ordered sequence of subproblems. Do not solve the subproblems and do not state a final answer. If the problem is already atomic, emit one subproblem rather than inventing unnecessary steps. Output only a numbered list. Task hints: SQuAD may locate evidence, resolve references, and identify the shortest supported span; HARP preserves answer choices and logical or mathematical constraints; MMLU preserves answer choices and the knowledge or reasoning needed to distinguish them; E2H-AMC preserves all mathematical conditions and the requested quantity.}
\promptrow{Least-to-most system}{Solve the supplied problem by answering the proposed subproblems in order. Treat the decomposition as a fallible plan: repair it if needed, preserve the original task, and verify the final result against the original question, followed by the applicable contract.}
\promptrow{Least-to-most user}{Original problem: \texttt{<problem>} Proposed decomposition (a plan, not authoritative evidence): \texttt{<decomposition>} Solve the original problem using this plan.}
\bottomrule
\end{tabular}
\caption{Prompts used by the adaptive-inference experiments.}
\label{tab:adaptive-prompts}
\end{table*}

The standard primary panels used temperature $0$, requested top-20 log-probabilities,
and used an 8,192-token output cap; repaired 70B panels used the larger cap recorded in
their run configuration. Generation used temperature $0.7$, equivalence checks used
temperature $0$, and wider-domain answer models used a 1,024-token cap.

\subsection{Signal computation and aggregation}
\label{app:signal-computation}
For each focal model, the difficulty target is the error rate of the remaining reference models. Leave-one-model-out construction prevents the focal model's own correctness from entering its target directly. This is important for the correctness-conditioned analyses, where including the focal model could create a mechanical association.

\paragraph{Perturbation inconsistency.}
For each item, the focal model answers the seven meaning-preserving variants. The extractor strips whitespace and superficial wrappers such as outer \texttt{\$} and \texttt{\textbackslash{}boxed\{\}}, removes commas and a trailing full stop, and maps empty,
``unfinished'', ``none'', ``nan'', and ``null'' outputs to missing. If $a_i$ is the normalised answer for variant $i$, we use the PC definition given in Eq.~\ref{eq:pc-definition} and report its complement $S_{\mathrm{pert}}(m,p)=1-\pc(m,p)$. The denominator is always $\binom{7}{2}=21$, including when an answer is missing. Thus
two ``A'' answers, three ``B'' answers, and two missing answers give
$\pc=4/21$ and $S_{\mathrm{pert}}=17/21$. In general, larger
$S_{\mathrm{pert}}$ means that the same model is less stable when the wording changes.

\paragraph{Majority-pair perturbation inconsistency.}
This uses the same seven normalised answers but retains in the numerator only pairs inside the largest answer group. If $n_{\max}$ is the largest count amongst the valid answers, the signal is given as 
\begin{equation}
  S_{\mathrm{maj}}(m,p)=1-\frac{\binom{n_{\max}}{2}}{\binom{7}{2}}.
\end{equation}

The fixed denominator means that missing answers and minority groups do not contribute matching pairs. Suppose there are three distinct responses from the seven perturbed inputs, and their counts are $(2, 3, 2)$ then $S_{\mathrm{maj}}=1-\binom{3}{2}/21=16/21$.

\paragraph{Semantic answer entropy.}
Seven additional answers to the unchanged original question are sampled at temperature $0.7$. Missing samples are discarded. Answers are processed in their stored order: an answer joins the first existing cluster whose representative is an exact string match or is judged equivalent by the temperature-zero answer-equivalence call; otherwise it starts a new cluster. If the resulting clusters have counts $n_1,\ldots,n_C$ and
$N=\sum_c n_c$, the signal is
\begin{equation}
  S_{\mathrm{sem}}(m,p)=-\sum_{c=1}^{C}\frac{n_c}{N}\log\!\left(\frac{n_c}{N}\right).
\end{equation}
For example, four samples in one semantic cluster and three in another produce $-\frac47\log\frac47-\frac37\log\frac37$; equivalent textual answers count as one cluster. Entropy is $0$ when all valid samples share one cluster and increases as the
sample mass spreads across distinct answers. An item with no valid sample receives a missing value.

\paragraph{Self-rated difficulty.}
The model is asked to rate the original problem for a strong AI model on an integer scale from 1 (trivial) to 5 (competition or olympiad level), at temperature $0$. Up to three prompt forms are tried until one parses as a single digit or a labelled
``rating''/``difficulty'' value. The returned integer is used directly: $S_{\mathrm{self}}(m,p)\in\{1,2,3,4,5\}$. Thus a parsed rating of 4 remains 4 and is oriented as greater predicted difficulty; this is a difficulty report, not a calibrated correctness.

\paragraph{Prompted numerical uncertainty.}
The model is asked for its probability of answering the original problem correctly as
a numeric value in $[0,100]$. The parser accepts a bare value or a value
labelled ``confidence'', ``probability'', or ``estimate'', and the first valid retry is
kept. To orient the signal towards uncertainty, a parsed confidence $q$ is stored as
\begin{equation}
  S_{\mathrm{num}}(m,p)=1-\frac{q}{100}.
\end{equation}
For example, $q=80$ becomes $0.20$, while $q=0$ becomes $1.00$.

\paragraph{Verbalised uncertainty.}
The model is asked for one of five labels: ``very sure'', ``somewhat sure'', ``uncertain'', ``somewhat unsure'', or ``very unsure''. The parser also accepts the same label after ``Certainty:'' or ``Confidence:''. These labels are encoded as $0,0.25,0.5,0.75,1$, respectively, so that ``very unsure'' has the largest value and ``very sure'' has the smallest. The first valid result among the three deterministic prompt forms is retained.

\paragraph{Mean token entropy.}
For each generated token $t$, the run stores the returned top-logprob candidates $\ell_{t1},\ldots,\ell_{tJ_t}$. Let $T'$ be the set of generated-token positions for which at least one top-logprob candidate is returned. The implementation exponentiates and renormalises only these returned candidates,
\begin{equation}
  \begin{aligned}
    \tilde p_{tj}    & =\frac{\exp(\ell_{tj})}{\sum_{r=1}^{J_t}\exp(\ell_{tr})}, \\
    H_t              & =-\sum_{j=1}^{J_t}\tilde p_{tj}\log\tilde p_{tj},         \\
    S_{\mathrm{ent}} & =\frac{1}{|T'|}\sum_{t\in T'}H_t.
  \end{aligned}
\end{equation}
The per-response value is then averaged over the available seven perturbation responses; the original-response analysis uses its single canonical response. This is generation-wide, so reasoning and formatting tokens are included. The interpretation is simple: a flatter returned next-token distribution gives larger $S_{\mathrm{ent}}$.

\paragraph{Least confidence.}
Using the same set $T'$ of token positions defined above, the implementation takes the probability of the highest-probability candidate at each position, $c_t=\exp(\ell_{t1})$, and averages it over those positions. It stores its complement:
\begin{equation}
  S_{\mathrm{LC}}=1-\frac{1}{|T'|}\sum_{t\in T'}c_t.
\end{equation}
Unlike the entropy calculation, this top-token probability is not renormalised over the returned candidate list. The item-level variant score is the mean of the available response-level scores. Thus a low average probability for the model's preferred next token gives a high least-confidence value.

\paragraph{Margin uncertainty.}
Let $T''$ be the set of generated-token positions for which at least two top-logprob candidates are returned. At each position in $T''$, the implementation computes the raw top-two probability gap $g_t=\exp(\ell_{t1})-\exp(\ell_{t2})$ and stores its complement averaged over these positions:
\begin{equation}
  S_{\mathrm{margin}}=1-\frac{1}{|T''|}\sum_{t\in T''}g_t.
\end{equation}
Small gaps therefore produce high uncertainty. As with least confidence, the response-level values are averaged across available perturbation responses, and the calculation is generation-wide. In plain terms, the signal is high when the two most likely next tokens are difficult to distinguish.

\paragraph{Solution length.}
Solution length is the completion-token count returned by the generation API, not a re-tokenisation of the stored text. For a canonical response, $S_{\mathrm{len}}(m,p)=T(m,p)$. For the perturbation-based item score, the code averages the available completion-token counts across the seven responses; for example, counts of 100, 120, and 140 give a score of 120. Missing counts are omitted from this arithmetic mean. Longer values therefore indicate more generated work or verbosity, but do not by themselves establish that the model was uncertain.

All signals are oriented so that larger values indicate greater uncertainty or predicted difficulty. We use Spearman correlation because the target is an ordinal ranking and a linear relationship is not required. Available model--dataset configurations are weighted equally in aggregate tables, preventing models with more complete auxiliary signals from dominating the summary.

\subsection{Sampling and uncertainty estimation}
\label{app:bootstrap}

For the RQ1 correlations, each bootstrap replicate samples the 500 item IDs with
replacement, recomputes Spearman's $\rho$ on the resampled pairs, and uses the 2.5th and
97.5th percentiles of 1,000 replicates as the reported 95\% interval. The
correctness-conditioned analyses use the observed correctness strata and fixed signal
quartile boundaries; resampling is clustered by problem ID within dataset because the
same problem is answered by multiple models. Panel-level correlations are computed
before equal-weight averaging, matching the estimand used in the main tables. Clustering
and fixed boundaries keep the intervals focused on item-sampling variation rather than
quietly changing the analysis definition within each replicate.

\subsection{Routing and adaptive-inference protocols}
\label{app:protocols}

The routing study uses 30 calibration and 70 test items per dataset for each weak model
(280 test items in total). Thresholds are selected separately by dataset and weak model
on the calibration partition and then frozen. The interval policy routes items whose
perturbation-consistency score lies between two thresholds; one-sided consistency,
solution-length, mean-token-entropy, and random policies are comparison controls. We use
an interval because the diagnostic band analysis below shows that escalation can be
least useful at the extreme-hard tail, where both models often fail, and at the easiest
tail, where escalation can replace a correct weak answer with a different answer.

The adaptive-inference study uses 400 calibration, 400 validation, and 1,200 test items
per weak model. Its chain-of-thought stage produces a checked solution; self-refinement
passes that solution to a second call as an explicitly untrusted draft; and
least-to-most first requests a minimal ordered decomposition before solving the
subproblems. All strategies use a visible task-specific ``Final answer'' contract.
Thresholds are fitted without test leakage: calibration constructs candidates, validation
selects the policy under the token budget, and the frozen policy is evaluated once on
the test split. Qwen3-8B uses native thinking with temperature $0.6$, top-$p=0.95$, and
top-$k=20$; Phi-4-14B uses deterministic decoding. The stage token ceilings and model
context sizes are recorded in the run configurations because they differ by model.

\section{Additional Results}
\label{app:additional-results}

This section reports supplementary measurements and explains what they add to the main
claims. The values are not intended as a second primary ranking of signals; coverage
differs across auxiliary runs and several comparisons are descriptive.

\subsection{Original-response coverage and truncation}
\label{app:original-responses}

An ``original-response configuration'' is one model--dataset panel: one of the eight
reference models answering the 500 canonical items in either HARP or E2H-AMC under the
original prompt. The sixteen panels therefore contain $16\times500=8{,}000$ original
responses. All rows had a valid extracted prediction and were retained. For compactness,
``DeepSeek-R1-70B'' in the table denotes DeepSeek-R1-\allowbreak Distill-\allowbreak Llama-70B.

\begin{table}[H]
  \centering
  \scriptsize
  \begin{tabular}{@{}llrr@{}}
    \toprule
    Dataset & Model             & Items & Length \\
    \midrule
    HARP    & DeepSeek-R1-70B   & 500   & 0      \\
    HARP    & Gemma-3-12B       & 500   & 9      \\
    HARP    & Gemma-3-27B       & 500   & 12     \\
    HARP    & Llama-3.3-70B     & 500   & 2      \\
    HARP    & Mistral-Small-24B & 500   & 10     \\
    HARP    & OLMo-2-13B        & 500   & 7      \\
    HARP    & Phi-4-14B         & 500   & 5      \\
    HARP    & Qwen3-8B          & 500   & 10     \\
    \midrule
    E2H-AMC & DeepSeek-R1-70B   & 500   & 37     \\
    E2H-AMC & Gemma-3-12B       & 500   & 4      \\
    E2H-AMC & Gemma-3-27B       & 500   & 9      \\
    E2H-AMC & Llama-3.3-70B     & 500   & 0      \\
    E2H-AMC & Mistral-Small-24B & 500   & 7      \\
    E2H-AMC & OLMo-2-13B        & 500   & 7      \\
    E2H-AMC & Phi-4-14B         & 500   & 4      \\
    E2H-AMC & Qwen3-8B          & 500   & 41     \\
    \bottomrule
  \end{tabular}
  \caption{Original-response panels used by the confidence--difficulty analysis. ``Length'' is the number of rows with a recorded output-token-limit finish reason.}
  \label{tab:original-response-panels}
\end{table}

The largest truncation counts occur for Qwen3-8B and DeepSeek-R1-70B on E2H-AMC, but
they remain a small fraction of their 500-item panels. Because the extraction procedure
recovered a valid answer from every retained row, removing these examples would
preferentially remove long reasoning traces and could bias solution-length and token
signals. We therefore report the counts transparently and retain the valid predictions;
the counts also motivate the separate output-cap documentation in Appendix~\ref{app:prompts}.

\subsection{Per-model signal correlations}
\label{app:per-model}

Table~\ref{tab:per-model-correlations} gives the per-model Spearman correlations for
the four primary signals, restricted to model--dataset configurations with complete
coverage.

\begin{table*}[t]
  \centering
  \scriptsize
  \begin{tabular}{@{}llrrrr@{}}
    \toprule
    Dataset & Model         & $1-\pc$ & Sem. entropy & Sol. length & Mean tok. entropy \\
    \midrule
    HARP    & Gemma-3-12B   & 0.585   & 0.529        & 0.635       & 0.253             \\
    HARP    & Gemma-3-27B   & 0.511   & 0.465        & 0.647       & 0.253             \\
    HARP    & Phi-4-14B     & 0.560   & 0.577        & 0.566       & 0.027             \\
    HARP    & Qwen3-8B      & 0.474   & 0.463        & 0.645       & 0.252             \\
    \midrule
    E2H-AMC & Gemma-3-12B   & 0.582   & 0.478        & 0.488       & 0.250             \\
    E2H-AMC & Gemma-3-27B   & 0.583   & 0.539        & 0.561       & 0.196             \\
    E2H-AMC & Llama-3.3-70B & 0.489   & 0.501        & 0.426       & 0.328             \\
    E2H-AMC & Phi-4-14B     & 0.573   & 0.515        & 0.459       & 0.155             \\
    \bottomrule
  \end{tabular}
  \caption{Per-model Spearman correlations with population difficulty for the four primary signals.}
  \label{tab:per-model-correlations}
\end{table*}

Perturbation inconsistency is positive in every complete panel, ranging from $0.474$ to
$0.585$ on HARP and from $0.489$ to $0.583$ on E2H-AMC. This consistency across focal
models supports the claim that the association is not caused by one model. Solution
length is often the strongest HARP signal, whereas token entropy is weaker and more
variable; this is why the paper treats generated effort as a useful but conceptually
distinct comparator rather than as confidence itself.

\subsection{Additional signal comparisons}
\label{app:additional-signal-comparisons}

Table~\ref{tab:additional-signal-correlations} reports equal-weight means over available
runs, including the primary signals and the auxiliary prompted, perturbation, token-level,
and answer-level variants.

\begin{table}[t]
  \centering
  \scriptsize
  \begin{tabular}{@{}p{0.43\columnwidth}rrr@{}}
    \toprule
    Signal                         & HARP  & E2H-AMC & Runs \\
    \midrule
    Majority-pair inconsistency    & 0.528 & 0.555   & 4    \\
    Perturbation inconsistency     & 0.532 & 0.557   & 4    \\
    Semantic entropy               & 0.509 & 0.508   & 4    \\
    Original solution length       & 0.573 & 0.489   & 8    \\
    Variant solution length        & 0.623 & 0.484   & 4    \\
    Original mean entropy          & 0.236 & 0.262   & 8    \\
    Variant mean entropy           & 0.289 & 0.276   & 4    \\
    Original least confidence      & 0.215 & 0.235   & 8    \\
    Variant least confidence       & 0.276 & 0.256   & 4    \\
    Original margin uncertainty    & 0.196 & 0.213   & 8    \\
    Variant margin uncertainty     & 0.246 & 0.232   & 4    \\
    Self-rated difficulty          & 0.466 & 0.390   & 5/4  \\
    Prompted numerical uncertainty & 0.386 & 0.293   & 5/4  \\
    Verbalised uncertainty         & 0.291 & 0.338   & 5/4  \\
    \bottomrule
  \end{tabular}
  \caption{Additional signal correlations. The run count is shown as HARP/E2H-AMC when coverage differs.}
  \label{tab:additional-signal-correlations}
\end{table}

The table shows why the main comparison should not be read as a universal ordering of
confidence measures. Variant solution length is strongest on HARP ($0.623$), while the
prompted measures are moderate and token-level measures are consistently weaker. The
similarity between the two perturbation definitions indicates that the main result is
not sensitive to whether all answer groups or only the largest group define agreement.
At the same time, the lower E2H-AMC correlations for several auxiliary signals show that
their usefulness depends on the task and on whether the signal is measured on the
original or perturbed responses.

For the correctness-conditioned aggregate, among correctly answered items, $1-\pc$ is
0.423 on HARP, 0.550 on E2H-AMC, and 0.486 when pooled; semantic entropy is 0.437 when
pooled; solution length is 0.535 when pooled; and mean token entropy is 0.142 when
pooled. Among incorrectly answered items, the corresponding pooled values are 0.179,
0.138, 0.211, and 0.164. The fixed quartile hard-minus-easy standardized effects for
correct items are 0.894 for $1-\pc$, 0.856 for semantic entropy, and 0.428 for mean
token entropy. Thus the positive associations are not explained solely by separating
correct from incorrect answers, although they are weaker within the incorrect stratum.

\subsection{Population-split and wider-domain checks}
\label{app:population-splits}

\begin{figure*}[!t]
  \centering
  \includegraphics[width=0.98\textwidth]{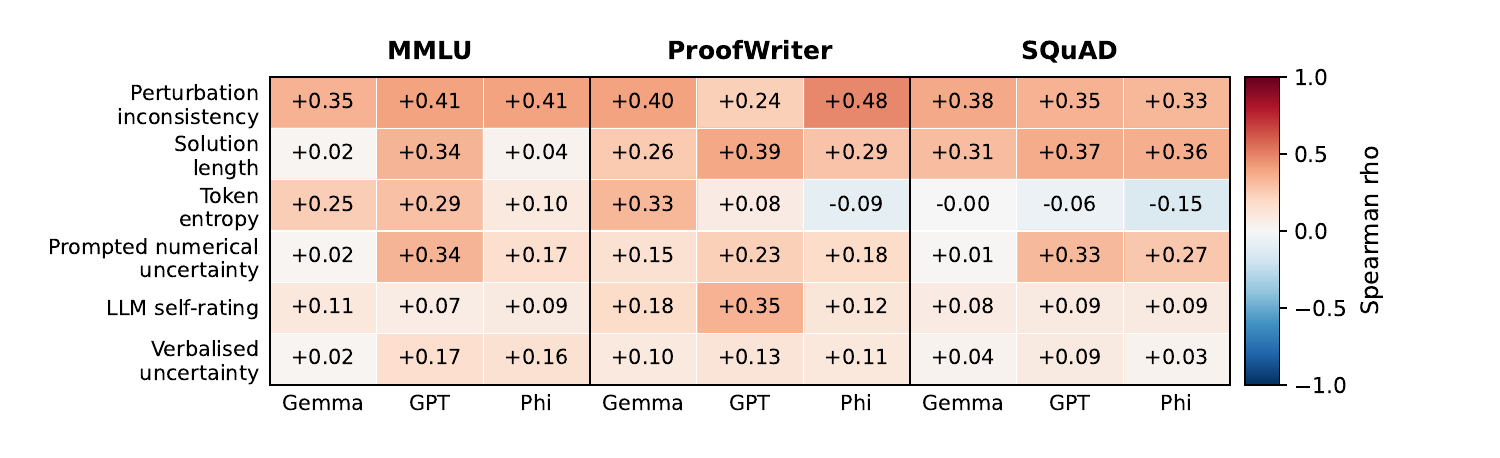}
  \caption{Wider-domain signal associations with model-population difficulty. Rows are signals; columns are the three focal models within each task.}
  \label{fig:wider-domain-heatmap}
\end{figure*}

For the primary four-versus-four population split, the analysis reports Spearman
correlations of 0.519 on HARP and 0.849 on E2H-AMC, Kendall correlations of 0.458 and
0.767, mean absolute difficulty differences of 0.163 and 0.147, and exact agreement
rates of 0.456 and 0.554, respectively. Across all 35 balanced four-versus-four splits,
the Spearman range is 0.371--0.528 on HARP (median 0.463) and 0.684--0.874 on E2H-AMC
(median 0.818). The stronger AMC agreement suggests that the operational difficulty
ranking is more stable there, while the wider HARP range confirms that reference-model
composition remains a real source of uncertainty.

Across the wider-domain configurations, perturbation inconsistency remains positive in
all nine cells, with correlations from 0.24 to 0.48. Solution length is also positive
throughout, whereas token entropy changes sign for several ProofWriter and SQuAD
configurations. Prompted measures are positive but smaller and less uniform. This
pattern supports cross-task generalization of the perturbation result while motivating
the paper's cautious interpretation of token-level and prompted signals.

\subsection{Routing diagnostics and results}
\label{app:routing}

Table~\ref{tab:routing-bands} reports the accuracy gain by diagnostic difficulty band.
The bands are used only to analyze where escalation helps; they are not test-time
correctness labels available to the router. The local-model column is the arithmetic mean
of Gemma-3-12B and Phi-4-14B accuracy.

\begin{table}[H]
  \centering
  \scriptsize
  \begin{tabular}{@{}llrrrr@{}}
    \toprule
    Dataset & Band    & $n$ & GPT-5.4 & Gemma/Phi avg. & Gain   \\
    \midrule
    E2H-AMC & Hardest & 33  & 0.333   & 0.000          & +0.333 \\
    E2H-AMC & Hard    & 35  & 0.686   & 0.229          & +0.457 \\
    E2H-AMC & Easy    & 37  & 0.919   & 0.608          & +0.311 \\
    E2H-AMC & Easiest & 95  & 0.937   & 1.000          & -0.063 \\
    MMLU    & Hardest & 15  & 0.467   & 0.000          & +0.467 \\
    MMLU    & Hard    & 15  & 0.867   & 0.100          & +0.767 \\
    MMLU    & Easy    & 49  & 0.918   & 0.520          & +0.398 \\
    MMLU    & Easiest & 121 & 0.983   & 1.000          & -0.017 \\
    \bottomrule
  \end{tabular}
  \caption{Accuracy and stronger-minus-local gain by diagnostic difficulty band.}
  \label{tab:routing-bands}
\end{table}

The gains peak in the intermediate bands and become negative in the easiest bands,
which provides the empirical rationale for an interval policy. The pattern also warns
against treating the least-consistent items as automatically valuable escalation targets:
some are beyond both models' capabilities. On the held-out medium-band evaluation,
escalation raises accuracy from 0.796 to 0.829 for Qwen3-8B and from 0.671 to 0.757 for
Phi-4-14B, while changing an incorrect weak answer to a correct answer on 3.9\% and
8.9\% of items, respectively. These are useful but modest gains, so the table supports
the routing hypothesis without establishing a generally superior policy.

\begin{table}[t]
  \centering
  \scriptsize
  \begin{tabular}{@{}lrrrrrr@{}}
    \toprule
    Model     & $n$ & Routed & Acc.  & Weak  & Gain   & Useful \\
    \midrule
    Qwen3-8B  & 280 & 92     & 0.829 & 0.796 & +0.032 & 0.039  \\
    Phi-4-14B & 280 & 92     & 0.757 & 0.671 & +0.086 & 0.089  \\
    \bottomrule
  \end{tabular}
  \caption{Medium-band routing results on the 280-item test partition. ``Useful'' is the fraction of items on which escalation changes an incorrect weak answer to a correct answer.}
  \label{tab:medium-band-routing}
\end{table}

\subsection{Adaptive-inference results}
\label{app:adaptive-inference}

Table~\ref{tab:adaptive-inference} reports the adaptive-inference test accuracy and
mean generated tokens on the test split.

\begin{table}[h]
  \centering
  \scriptsize
  \begin{tabular}{@{}llrr@{}}
    \toprule
    Model     & Strategy         & Accuracy & Mean tokens \\
    \midrule
    Qwen3-8B  & Adaptive policy  & 0.856    & 6,121       \\
    Qwen3-8B  & Chain-of-thought & 0.858    & 4,143       \\
    Qwen3-8B  & Self-refinement  & 0.867    & 8,922       \\
    Qwen3-8B  & Least-to-most    & 0.849    & 5,847       \\
    Phi-4-14B & Adaptive policy  & 0.728    & 1,359       \\
    Phi-4-14B & Chain-of-thought & 0.729    & 755         \\
    Phi-4-14B & Self-refinement  & 0.725    & 2,159       \\
    Phi-4-14B & Least-to-most    & 0.734    & 1,630       \\
    \bottomrule
  \end{tabular}
  \caption{Adaptive-inference test accuracy and mean generated tokens on the test split.}
  \label{tab:adaptive-inference}
\end{table}

The adaptive policy is close to chain-of-thought accuracy for both models, but it does
not exceed the best fixed strategy: self-refinement is strongest for Qwen3-8B, while
least-to-most is strongest for Phi-4-14B. Its token cost is between the cheaper and more
expensive fixed procedures. The result therefore demonstrates a cost--quality frontier
and the feasibility of signal-guided allocation, rather than universal accuracy gains
from adaptive inference.

\end{document}